\documentclass[pmlr]{jmlr}

\usepackage{longtable}

\usepackage{booktabs}

\theorembodyfont{\upshape}
\theoremheaderfont{\scshape}
\theorempostheader{:}
\theoremsep{\newline}

\jmlrvolume{}
\firstpageno{1}
\editors{Sophia Sanborn, Christian Shewmake, Simone Azeglio, Nina Miolane}

\jmlryear{2023}
\jmlrworkshop{Symmetry and Geometry in Neural Representations}

 \author{\Name{Donato Crisostomi} \Email{crisostomi@di.uniroma1.it}\\
 \addr Sapienza University of Rome
 \AND
 \Name{Irene Cannistraci} \Email{cannistraci@di.uniroma1.it}\\
 \addr Sapienza University of Rome
 \AND
 \Name{Luca Moschella} \Email{moschella@di.uniroma1.it}\\
 \addr Sapienza University of Rome
 \AND
  \Name{Pietro Barbiero} \Email{pb737@cam.ac.uk }\\
 \addr University of Cambridge
 \AND
  \Name{Marco Ciccone} \Email{marco.ciccone@polito.it}\\
 \addr Politecnico di Torino
 \AND
 \Name{Pietro Liò} \Email{pl219@cam.ac.uk } \\
 \addr University of Cambridge
 \AND
 \Name{Emanuele Rodolà} \Email{rodola@di.uniroma1.it} \\
 \addr Sapienza University of Rome
 }

\usepackage[utf8]{inputenc} 
\usepackage[T1]{fontenc}    
\usepackage{amsfonts}       
\usepackage{nicefrac}       
\usepackage{microtype}      
\usepackage{algorithmicx, algorithm}
\usepackage{algpseudocode}
\usepackage{caption}
\usepackage{bm}
\usepackage{multirow}
\usepackage{overpic}
\usepackage{colortbl}
\usepackage{wrapfig}

\usepackage{multirow}
\usepackage{multicol}
\usepackage{pgf}
\usepackage{pgfplots}
\usepackage{tikz-3dplot}
\pgfplotsset{compat=1.16}

\AtEndPreamble{
  \usepackage{cleveref}
}

\usepackage[acronym]{glossaries}
\glsdisablehyper
\newacronym{dl}{DL}{Deep Learning}
\newacronym{cv}{CV}{Computer Vision}
\newacronym{rr}{RR}{Relative Representations}
\newacronym{rlsa}{RLSA}{Relative Latent Space Aggregation}
\newacronym{pca}{PCA}{Principal Component Analysis}

\newacronym{nn}{NN}{Neural Network}
\newacronym{dnn}{DNN}{Deep Neural Network}
\newacronym{mlp}{MLP}{MultiLayer Perceptron}
\newacronym{cnn}{CNN}{Convolutional Neural Network}
\newacronym{ae}{AE}{AutoEncoder}
\newacronym{vae}{VAE}{Variational AutoEncoder}
\newacronym{vit}{ViT}{Vision Transformer}

\newacronym{mse}{MSE}{Mean Squared Error}
\newacronym{cka}{CKA}{Centered Kernel Alignment}
\newacronym{cca}{CCA}{Canonical Correlation Analysis}
\newacronym{svcca}{SVCCA}{Singular Value CCA}
\newacronym{pwcca}{PWCCA}{Projection Weighted CCA}
\newacronym{svd}{SVD}{Singular Value Decomposition}
\newacronym{hsic}{HSIC}{Hilbert-Schmidt Independence Criterion}

\newacronym{abs}{Abs.}{Absolute}
\newacronym{cos}{Cos.}{Cosine}

\newacronym{at}{AT}{Affine Transformation}
\newacronym{tr}{TR}{Translation}
\newacronym{lt}{LT}{Linear Transformation}
\newacronym{is}{IS}{Isotropic Scaling}
\newacronym{ot}{OT}{Orthogonal Transformation}
\newacronym{pt}{PT}{Permutation}

\newcommand{\cifar}{\texttt{CIFAR100}}
\newcommand{\tinyimagenet}{\texttt{TinyImageNet}}
\newcommand{\efficientnet}{\texttt{EfficientNet}}
\newcommand{\vanillacnn}{\texttt{VanillaCNN}}

\newcommand{\R}{\bm{R}}
\newcommand{\Rprime}{\R^\prime}

\newcommand{\Sm}{\bm{S}}
\newcommand{\Smp}{\bm{S'}}
\newcommand{\x}{\bm{x}}

\newcommand{\transp}{\mathsf{T}}
 
\title[From Charts to Atlas: Merging Latent Spaces into One]{From Charts to Atlas: Merging Latent Spaces into One}

\begin{document}
\pagenumbering{gobble}

\maketitle

\begin{abstract}
    Models trained on semantically related datasets and tasks exhibit comparable inter-sample relations within their latent spaces.
    We investigate in this study the aggregation of such latent spaces to create a unified space encompassing the combined information.
    To this end, we introduce \gls{rlsa}, a two-step approach that first renders the spaces comparable using relative representations, and then aggregates them via a simple mean.
    We carefully divide a classification problem into a series of learning tasks under three different settings: sharing samples, classes, or neither. We then train a model on each task and aggregate the resulting latent spaces. 
    We compare the aggregated space with that derived from an end-to-end model trained over all tasks and show that the two spaces are similar. We then observe that the aggregated space is better suited for classification, and empirically demonstrate that it is due to the unique imprints left by task-specific embedders within the representations.
    We finally test our framework in scenarios where no shared region exists and show that it can still be used to merge the spaces, albeit with diminished benefits over naive merging.
\end{abstract} 

\section{Introduction}
%
The success of neural networks can partly be attributed to the latent spaces they learn. In fact, given a sufficiently general task, the deeper layers of a network are able to learn semantically meaningful representations that capture the inherent data structure. If these representations solely relied on semantics, the latent space would exhibit invariance to the model's architecture and training process and be seamlessly transferable to other tasks sharing the same semantics. In practice, however, training stochasticities such as initialization and data shuffling hinder the immediate comparability of latent spaces.
Nevertheless, when spaces are rich in semantics, the distance between data points becomes indicative of their similarity. That is, samples located close to each other are likely to share similarities, and conversely, those farther apart are expected to be more distinct.
This intuitive notion suggests that the latent spaces of models trained on semantically analogous datasets and tasks relying on the same underlying structure should exhibit similar inter-sample relations.
Returning to a geometric perspective, this implies that the inter-sample distances should remain consistent across the two spaces. Building upon this insight, \cite{moschella2023relative} introduced relative representations as a way to compare latent spaces: when the similarities among samples are consistent across the two spaces, the latter become comparable.

%
In this paper, we investigate a natural follow-up question: when, and under what assumptions, \textit{can two spaces be merged into one?}
In principle, given two comparable representations that may partially overlap or be entirely disjoint, it should be possible to generate a unified representation in which both coexist consistently. We refer to this problem as \textit{Latent Space Aggregation}. Space aggregation raises several questions on i) the representational power of the unified representation space, ii) its ability to accommodate both spaces without collisions, and iii) its robustness to complementary information present in only one of the two spaces. In fact, naively aggregating the sample representations by computing their mean {\em in absolute coordinates} would not account for the different latent configurations caused by training stochasticities, resulting in an inconsistent aggregation of different entities based on spurious random factors. 
\begin{figure}
    \begin{center}
    \begin{overpic}[width=\linewidth]{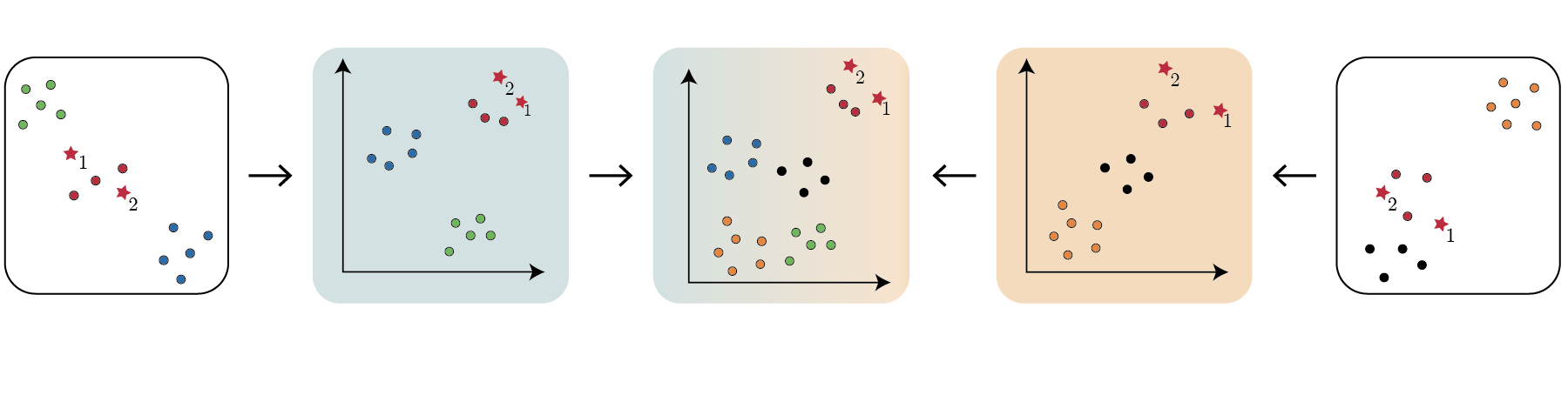}
        %
        \put(6.5,-1){$\mathcal{X}$}
        \put(91.5,-1){$\mathcal{Y}$}
        \put(27.5,-1){$\mathcal{X}_\text{rel}$}
        \put(71.5,-1){$\mathcal{Y}_\text{rel}$}
        \put(43,-1){RLSA($\mathcal{X}_\text{rel}$, $\mathcal{Y}_\text{rel}$)}

        \put(60, 23) {\small $s(\cdot,$ \includegraphics[height=5pt]{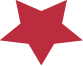}$_2)$}
        \put(37, 23){\small $s(\cdot,$ \includegraphics[height=5pt]{figures/star.png}$_2)$}
        \put(16, 23){\small $s(\cdot,$ \includegraphics[height=5pt]{figures/star.png}$_2)$}
        \put(75, 3.5){\small $s(\cdot,$ \includegraphics[height=5pt]{figures/star.png}$_1)$}
        \put(52, 3.5){\small $s(\cdot,$ \includegraphics[height=5pt]{figures/star.png}$_1)$}
        \put(31, 3.5){\small $s(\cdot,$ \includegraphics[height=5pt]{figures/star.png}$_1)$}
    \end{overpic}
    \vspace{0.2cm}
    \caption{\textbf{Framework description}. Given two absolute spaces $\mathcal{X}$ and $\mathcal{Y}$, we first project these spaces into two comparable relative representations $\mathcal{X}_\text{rel}$, $\mathcal{Y}_\text{rel}$. Then, we combine these representations into a single, unified relative space RLSA($\mathcal{X}_\text{rel}$, $\mathcal{Y}_\text{rel}$).}
    \vspace{-1cm}
    \label{fig:teaser}
    \end{center}
\end{figure}

%
Motivated by the above challenges, we propose \acrfull{rlsa}. 
Our approach involves two steps: we first switch to a relative representation where the latent spaces are represented with respect to a set of anchors, and then aggregate the obtained representations by computing their mean.
The first step makes the spaces comparable, enabling a meaningful aggregation of samples that are common to multiple latent spaces, at the same time avoiding collisions. 

%
To test our framework, we partition a classification dataset into multiple learning \textit{tasks}. These tasks can vary in terms of class composition, such as covering disjoint subsets of classes, or in sample composition, such as being sampled with different class distributions. These diverse tasks enable us to train task-specific models, extract their latent spaces, and subsequently examine their aggregation.
We consider three different cases: i) tasks sharing a set of samples, ii) tasks sharing the same classes but disjoint sample sets, and iii) tasks disjoint both at the class and at the sample level. 
In the first case, we select the anchors from the shared samples, while in the disjoint scenarios they are sampled from unseen samples in the training dataset.
We then analyze the quality of the aggregation by i) comparing it to the space of an end-to-end model trained on all the tasks, ii) assessing the performance of a classifier over the aggregated space, and iii) quantifying the separability of the classes within it. 
We show that the best results are obtained when sharing samples, while the benefits decrease in the disjoint scenarios. Finally, we release an extensible and modular codebase\footnote{\href{https://github.com/crisostomi/latent-aggregation}{https://github.com/crisostomi/latent-aggregation}} to foster reproducibility and further research in the problem.

To summarize, our contributions are three-fold:
\begin{enumerate}
    \item We propose for the first time a framework for latent space aggregation, merging different latent spaces without requiring weight averaging, sharing, or any model-specific details;
    \item We evaluate our framework on aggregating tasks sharing samples, classes, or neither, assessing representational power, class separability, and similarity to the global space; 
    \item We investigate the improved performance over class-disjoint tasks, empirically demonstrating that it is a natural consequence of utilizing task-specific embedders.
\end{enumerate}

\section{Related work}
Closest to our work are the fields of model merging and representational similarity analysis. In model merging, the goal is to combine multiple models into a single model. This is usually done by merging the models' parameters with some aggregation function, \emph{e.g.}, by averaging \citep{fedavg}. Most recently, Git Re-Basin \citep{Ainsworth2022-yu} leverages advances in linear mode connectivity \citep{Garipov2018-pz,Benton2021-pr,Frankle2019-gj} to first map the two models into the same basin, and then interpolate among them. It is important to note that, unlike model merging, \emph{our approach does not require any architectural details from the models} and is therefore applicable to a set of spaces originating from any set of models. On the other hand, our approach does not provide a trained model, but rather a unified space that can be used as a starting point to train other models.

In representational similarity analysis, the goal is to compare the representations of different models. Several measures have been proposed for the task \citep{sim-measure-survey,Shahbazi2021-qp,Raghu2017-gx,Tang2020-rq,Williams2021-ww}, with the most prominent being \gls{cka} \citep{Kornblith2019-vr}. In this work, we use \gls{cka} to compare the representations of the models in the aggregated space.

\section{Approach}

\paragraph{Relative representations}
We leverage relative representations \citep{moschella2023relative} to render the seemingly different latent spaces comparable. In practice, we start by selecting a subset $\mathbb{A}$ of the training data $\mathcal{X}$, denoted as anchor samples. Every sample in the training distribution will be represented as a function of the embedded anchors $\mathbf{a}_j = f_\Theta(a_j)$ with $a_j \in \mathbb{A}$. As a measure capturing the relation between the anchors and the other samples, we consider the cosine similarity $s: \mathbb{R}^d \times \mathbb{R}^d \mapsto \mathbb{R}$, yielding a scalar score $r \in [-1, 1]$ between two absolute representations $\mathbf{x}_i, \mathbf{x}_j$. The relative representation of $ x_i \in \mathcal{X}$ as a function of the anchors $\mathbb{A}$ is then given by
\begin{equation}
    r(x_i) = \left(%
    s(\mathbf{x}_i, \mathbf{a}_1), s(\mathbf{x}_i, \mathbf{a}_2), \dots, s(\mathbf{x}_i, \mathbf{a}_{\|\mathbb{A}\|})\,.%
    \right)
\end{equation}
In the following, we will call the embedding space the \emph{absolute space}, and the set of relative representations the \emph{relative space}. 

\paragraph{Space aggregation}
Given a set of $M$ relative spaces $X_r^{(1)}, \dots, X_r^{(M)}$, we can define an aggregation function that maps them to a single space. If a sample $x$ appears in $K \leq M$ spaces, we can assemble its relative representations $x_r^{(1)}, \dots, x_r^{(K)}$ just by taking their mean:
\begin{equation}
    x_\text{aggregated} = \frac{1}{K} \sum_{k=1}^K X_r^{(k)}\,.
\end{equation}
This trivially entails that when the spaces are disjoint, each sample representation will just be its relative representation in its original space. The aggregation accounts for noise in the relative representations, as the same samples may be represented differently in the different relative spaces. Nevertheless, in the optimal case in which they perfectly align, the relative representation will be equal for all the spaces and therefore still equal to the mean. Due to the different experimental settings, the anchor set is chosen differently for each scenario. In particular, when a portion of samples is shared, we select our anchors there. In fact, these are samples that the model can reliably embed in the latent space and can be used as reference points to give structure to the new aggregated space. Intuitively, if the resulting latent regions are comparable and shared, the out-of-distribution samples should be triangulated in a consistent manner.
On the other hand, when the tasks are disjoint, the anchors are extracted from a small subset of the global dataset that is not shown to the model. Indeed, if these were used for training, the task-specific models would be partially aware of the overall class distribution, while we enforce them to see only the samples and classes of the task they are trained on. Importantly, the projection anchors are the same across all the tasks. Within the anchor set, 
we follow \cite{moschella2023relative} and employ random sampling.


\section{Experiments}
We consider two datasets throughout our experiments: \texttt{CIFAR100} \citep{cifar100} and \texttt{TinyImagenet} \citep{tinyimagenet}. We refer to the latter as \texttt{TINY} in the tables for brevity. These offer a good trade-off between scale and tractability. In fact, the first experiment alone requires training and aggregating $70$ task-specific models.
Dataset details can be found in \cref{appendix:dataset-details}.
As for the models, we experiment with a standard convolutional neural network trained from scratch and a pre-trained \texttt{EfficientNet} due to its superior performance on \texttt{CIFAR100} \citep{tan2019efficientnet}.
Common to the experiments will be a quantitative measure of the similarity between the aggregated space and the space of a model trained end-to-end on the whole dataset. To this end, we employ \gls{cka} \citep{Kornblith2019-vr} as it is the most commonly used metric for comparing neural representations. We measure the separability of the considered spaces as the ratio of the inter-class distances over the average of the intra-class distances. The latter is described in detail in  \cref{appendix:separability}. For space reasons, we will consider only one of the datasets or a subset of the configurations when the results are consistent, reporting the remaining experiments in \cref{appendix:additional-exps}.
\pagebreak
\subsection{Aggregating tasks sharing samples and classes}\label{sec:exp-T1}
\paragraph{Setup}

\begin{wrapfigure}{r}{0.45\textwidth}
  \centering
  \includegraphics[width=0.44\textwidth]{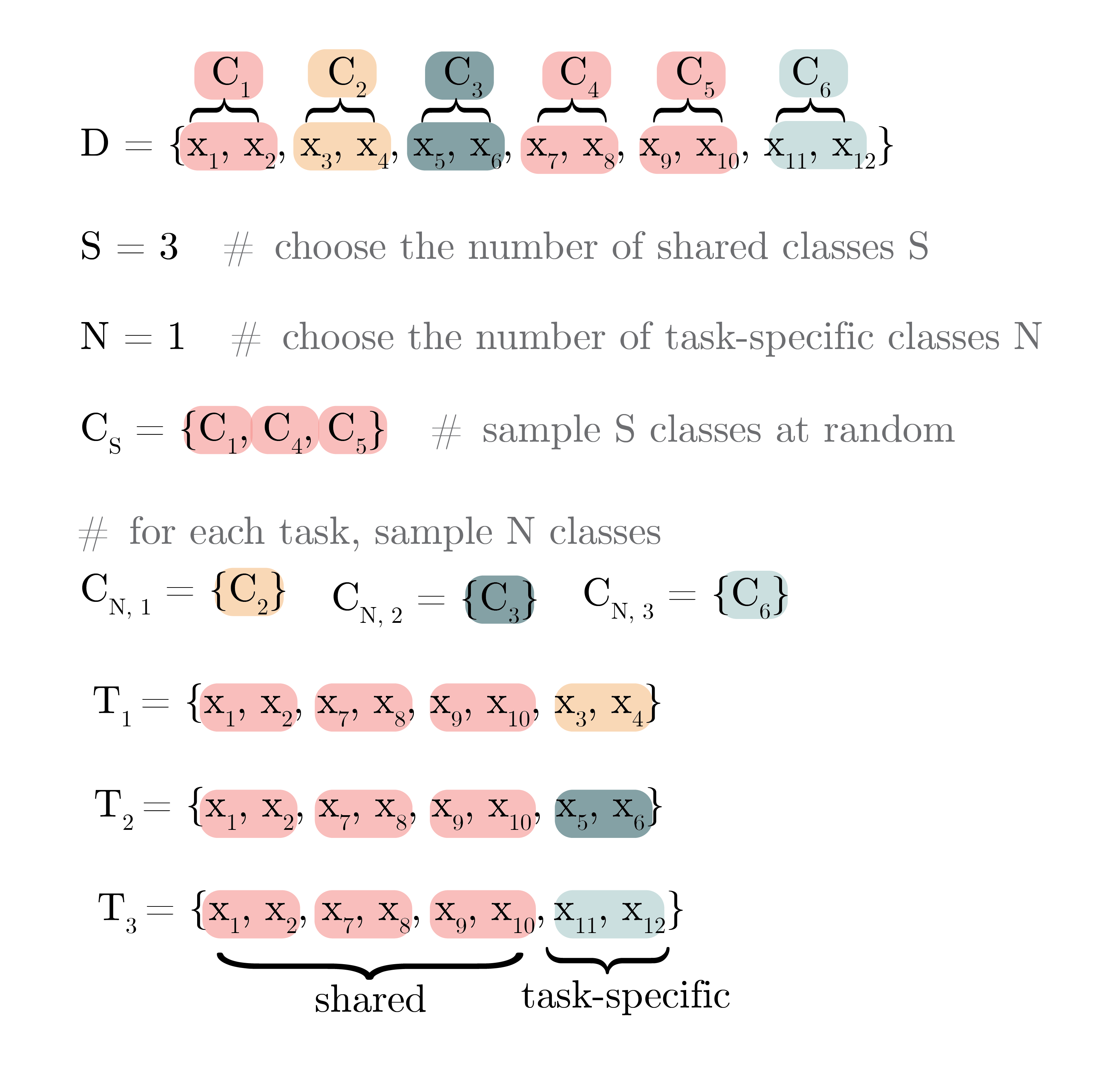} 
  \vspace{-0.5cm}
  \caption{Experiment outline.}\label{fig:exp1}
\end{wrapfigure}
In this experiment, we want to test the ability of our framework to merge latent spaces originating from models trained on partially disjoint tasks, \emph{i.e.} classification tasks that share a nucleus of classes but also have a set of task-specific classes. If a class is present in the task, either shared with the other tasks or not, all its samples are considered in the task. 
Given $C$ total classes in the dataset, we determine a number $S$ of classes that will be \emph{shared} across the tasks, and a number $N$ of classes that will be \emph{novel} for each task. 
This subdivision results in $K = \frac{(C-S)}{N}$ discrete tasks, each comprising all the $S$ shared classes and a set of $N$ task-specific novel classes. Subsequently, we train $K+1$ models: one inclusive of all the $C$ classes and one for each task. Each task-specific model is thus characterized by a latent space segregated into shared and novel samples.
The latent spaces defined by the $K$ task-specific models are then represented as similarities from a set of $256$ anchors sampled only from the shared samples. Intuitively, expecting the subspaces corresponding to the shared classes to be similar across the different models, these shared anchors should be able to impose a consistent structure on the merged space that will accommodate the novel samples.
Being represented in a possibly different way in each space, we obtain $K$ representations for each shared sample. The latter are averaged to get the final representation for the unified space. On the other hand, being only represented once across all the spaces, the task-specific novel samples can be added to the consolidated space without further ado.

\paragraph{Comparison with the original space}
As we can quantitatively assess in \Cref{tab:cka-part-shared-part-novel}, for each combination of $S$ and $N$, the space obtained by aggregating the latent spaces corresponding to the models trained on the partially disjoint tasks exhibits high similarity to the one obtained by training on the whole dataset. In fact, the values are always higher than $0.7$ and often higher than $0.8$. 
This can be also qualitatively appreciated in \Cref{fig:part-shared-part-novel-pca}.
As expected, the \gls{cka} among the samples from the shared classes is always higher than that of the samples from task-specific classes. Interestingly, the \gls{cka} is still high even when considering low values for $S$, \emph{i.e.} when having a small number of shared classes. This suggests that the framework can merge latent spaces even when the shared portion of samples is small.

\begin{table}
  \begin{center}
  \begin{small}
  \begin{tabular}{cccccccccc}
    \toprule
            &       &      &          & \multicolumn{6}{c}{CKA}                                                                                                                                                                   \\
    \cmidrule(lr){5-10}
    Dataset & S     & N    & \# tasks & non-shared                              & shared                                    & overall                               & non-shared & shared & overall                               \\
    \midrule
    \parbox[t]{2mm}{\multirow{10}{*}{ \rotatebox[origin=c]{90}{\texttt{CIFAR100}} }}
            &       &      &          & \multicolumn{3}{c}{\texttt{VanillaCNN}} & \multicolumn{3}{c}{\texttt{EfficientNet}}                                                                                                       \\    \cmidrule(lr){5-7} \cmidrule(lr){8-10}
            & $80$  & $10$ & 2        & 0.88                                    & 0.90                                      & \cellcolor[rgb]{0.74, 0.38, 0.37}0.90 & 0.85       & 0.90   & \cellcolor[rgb]{0.74, 0.39, 0.38}0.89 \\
            & $60$  & $10$ & 4        & 0.85                                    & 0.92                                      & \cellcolor[rgb]{0.74, 0.39, 0.38}0.89 & 0.82       & 0.88   & \cellcolor[rgb]{0.76, 0.44, 0.43}0.85 \\
            & $40$  & $10$ & 6        & 0.83                                    & 0.91                                      & \cellcolor[rgb]{0.76, 0.43, 0.43}0.86 & 0.80       & 0.89   & \cellcolor[rgb]{0.78, 0.47, 0.47}0.83 \\
            & $20$  & $10$ & 8        & 0.74                                    & 0.86                                      & \cellcolor[rgb]{0.82, 0.58, 0.57}0.76 & 0.77       & 0.87   & \cellcolor[rgb]{0.8, 0.53, 0.52}0.79  \\
    \cmidrule(lr){5-7} \cmidrule(lr){8-10}
            & $80$  & $5$  & 4        & 0.90                                    & 0.93                                      & \cellcolor[rgb]{0.72, 0.34, 0.34}0.92 & 0.81       & 0.90   & \cellcolor[rgb]{0.75, 0.4, 0.39}0.88  \\
            & $60$  & $5$  & 8        & 0.82                                    & 0.90                                      & \cellcolor[rgb]{0.75, 0.42, 0.41}0.87 & 0.83       & 0.87   & \cellcolor[rgb]{0.77, 0.45, 0.44}0.84 \\
            & $40$  & $5$  & 12       & 0.81                                    & 0.87                                      & \cellcolor[rgb]{0.78, 0.47, 0.46}0.84 & 0.78       & 0.89   & \cellcolor[rgb]{0.78, 0.49, 0.48}0.82 \\
            & $20$  & $5$  & 16       & 0.78                                    & 0.87                                      & \cellcolor[rgb]{0.8, 0.53, 0.52}0.80  & 0.77       & 0.88   & \cellcolor[rgb]{0.8, 0.53, 0.52}0.79  \\
    \midrule
    \parbox[t]{2mm}{\multirow{3}{*}{ \rotatebox[origin=c]{90}{\texttt{TINY}} }}
            &       &      &          & \multicolumn{3}{c}{\texttt{VanillaCNN}} & \multicolumn{3}{c}{\texttt{EfficientNet}}                                                                                                       \\    \cmidrule(lr){5-7} \cmidrule(lr){8-10}
            & $100$ & $25$ & 4        & 0.85                                    & 0.91                                      & \cellcolor[rgb]{0.74, 0.39, 0.39}0.88 & 0.72       & 0.81   & \cellcolor[rgb]{0.82, 0.59, 0.58}0.76 \\
            & $50$  & $25$ & 6        & 0.83                                    & 0.91                                      & \cellcolor[rgb]{0.77, 0.45, 0.44}0.85 & 0.67       & 0.80   & \cellcolor[rgb]{0.86, 0.66, 0.65}0.71 \\
    \bottomrule
  \end{tabular}
  \end{small}
  \end{center}
  \vspace{-0.5cm}
  \caption{
  \texttt{Experiment 1}. Comparison of the aggregated spaces with the space of an end-to-end model trained on the whole dataset. $S$ and $N$ represent the number of shared and novel classes per task out of a total of $C$ classes. We compute \gls{cka} both for the \emph{overall} space and for the subspaces defined by \emph{non-shared} and \emph{shared} classes.
  }
  \label{tab:cka-part-shared-part-novel}
\end{table}

\paragraph{Representational power of the aggregated space}
We also want to test the representational power of the aggregated space. To this end, we train a simple classifier on the aggregated space and compare its performance with the end-to-end trained classifier.
\Cref{tab:part-shared-part-novel-accuracy} shows that the former obtains better performance than one trained over the original space and, most importantly, that the performance increases when decreasing the number of shared classes. While this is a surprising result, we will see in \cref{sec:exp-T3} that it is due to task-specific embedders.
The gap in accuracy is impressive, with $+33$, $+20$ accuracy points in \cifar{} using \vanillacnn{} and \efficientnet{} and a $+14$, $+8$ increase for \tinyimagenet{}. To verify that it is enough to merge the spaces without first passing to a unified representation, we also test with a classifier trained on the union of the absolute embedding spaces. As we can see in \cref{tab:part-shared-part-novel-accuracy}, the performance is way lower than that of the classifier trained on the aggregated space, confirming that the relative representation is beneficial.

\begin{table}
\begin{center}
  \resizebox{\textwidth}{!}{%
    \begin{tabular}{cccccccccccccc} %

      \toprule
      Dataset & $S$   & $N$  & tasks & vanilla                                 & non-shared & shared                                    & total & improv                                 & vanilla & non-shared & shared & total & improv                                 \\
      \midrule
      \parbox[t]{2mm}{\multirow{9}{*}{ \rotatebox[origin=c]{90}{\texttt{CIFAR100}} }}
              &       &      &       & \multicolumn{4}{c}{\texttt{VanillaCNN}} & 0.39       & \multicolumn{4}{c}{\texttt{EfficientNet}} & 0.70                                                                                                                            \\    \cmidrule(lr){5-9} \cmidrule(lr){10-14}
              & $80$  & $10$ & 2     & 0.36                                    & 0.60       & 0.39                                      & 0.43  & \cellcolor[rgb]{0.89, 0.94, 0.91}+0.04 & 0.68    & 0.80       & 0.71   & 0.73  & \cellcolor[rgb]{0.9, 0.94, 0.92}+0.02  \\
              & $60$  & $10$ & 4     & 0.39                                    & 0.64       & 0.45                                      & 0.53  & \cellcolor[rgb]{0.82, 0.9, 0.85}+0.14  & 0.72    & 0.82       & 0.76   & 0.79  & \cellcolor[rgb]{0.86, 0.92, 0.88}+0.08 \\
              & $40$  & $10$ & 6     & 0.42                                    & 0.64       & 0.50                                      & 0.58  & \cellcolor[rgb]{0.78, 0.87, 0.82}+0.19 & 0.75    & 0.87       & 0.80   & 0.84  & \cellcolor[rgb]{0.82, 0.9, 0.85}+0.14  \\
              & $20$  & $10$ & 8     & 0.47                                    & 0.65       & 0.52                                      & 0.62  & \cellcolor[rgb]{0.75, 0.86, 0.8}+0.23  & 0.80    & 0.88       & 0.84   & 0.87  & \cellcolor[rgb]{0.8, 0.88, 0.83}+0.17  \\
      \cmidrule(lr){2-14}
              & $80$  & $5$  & 4     & 0.37                                    & 0.77       & 0.41                                      & 0.49  & \cellcolor[rgb]{0.85, 0.91, 0.87}+0.10 & 0.71    & 0.84       & 0.72   & 0.75  & \cellcolor[rgb]{0.89, 0.93, 0.9}+0.05  \\
              & $60$  & $5$  & 8     & 0.39                                    & 0.71       & 0.45                                      & 0.55  & \cellcolor[rgb]{0.8, 0.89, 0.84}+0.16  & 0.76    & 0.85       & 0.78   & 0.81  & \cellcolor[rgb]{0.85, 0.91, 0.87}+0.11 \\
              & $40$  & $5$  & 12    & 0.44                                    & 0.74       & 0.49                                      & 0.64  & \cellcolor[rgb]{0.74, 0.85, 0.78}+0.25 & 0.80    & 0.90       & 0.80   & 0.86  & \cellcolor[rgb]{0.81, 0.89, 0.84}+0.16 \\
              & $20$  & $5$  & 16    & 0.51                                    & 0.76       & 0.55                                      & 0.72  & \cellcolor[rgb]{0.68, 0.82, 0.74}+0.33 & 0.83    & 0.93       & 0.83   & 0.90  & \cellcolor[rgb]{0.77, 0.87, 0.81}+0.20 \\
      \cmidrule(lr){2-14}
      \parbox[t]{2mm}{\multirow{3}{*}{ \rotatebox[origin=c]{90}{\texttt{TINY}} }}
              &       &      &       & \multicolumn{4}{c}{\texttt{VanillaCNN}} & 0.22       & \multicolumn{4}{c}{\texttt{EfficientNet}} & 0.69                                                                                                                            \\    \cmidrule(lr){5-9} \cmidrule(lr){10-14}
              & $100$ & $25$ & 4     & 0.22                                    & 0.37       & 0.23                                      & 0.30  & \cellcolor[rgb]{0.87, 0.92, 0.89}+0.08 & 0.68    & 0.75       & 0.71   & 0.73  & \cellcolor[rgb]{0.89, 0.93, 0.9}+0.05  \\
              & $50$  & $25$ & 6     & 0.24                                    & 0.36       & 0.36                                      & 0.36  & \cellcolor[rgb]{0.82, 0.9, 0.85}+0.14  & 0.72    & 0.77       & 0.74   & 0.77  & \cellcolor[rgb]{0.86, 0.92, 0.88}+0.08 \\
      \bottomrule                                                                                                                                                                                                                                                      \\
    \end{tabular}
    }
    \end{center}
    \vspace{-0.8cm}
  \caption{
  Classification accuracy comparison. Each quarter shows a dataset-model combination, with end-to-end model accuracy on the right. For each $S$, $N$ combination, we report the accuracy of a classifier trained on the aggregated space, along with accuracy when considering only \emph{non-shared} and \emph{shared} classes. \emph{Improv} is the improvement over the end-to-end model, while \emph{vanilla} the accuracy of naive merging.
  }
  \label{tab:part-shared-part-novel-accuracy}
\end{table}

\begin{figure}
  \centering
  \includegraphics[width=\textwidth]{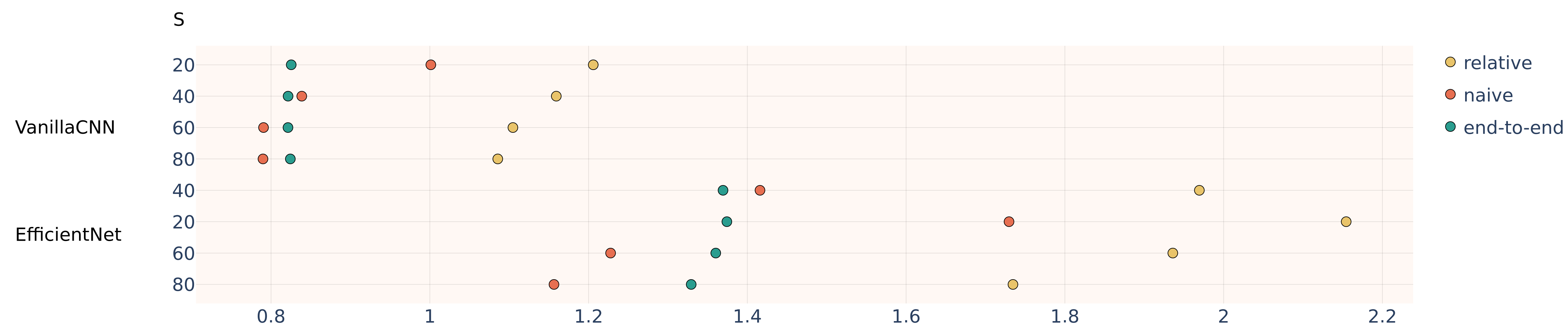}
    \vspace{-0.6cm}
  \caption{
  Separability analysis over \texttt{CIFAR100}: we compare the mean separability of the resulting space when using relative and naive aggregation for different values of $S$ and $N=10$, as well as the separability over the space of the end-to-end model trained over the whole dataset.
  The relatively aggregated space consistently results in the highest separability.
  }
  \vspace{-0.5cm}
  \label{fig:exp1-separability}
\end{figure}

To check whether different non-shared classes end up in the same region in the aggregated space, we measure the mean separability for any pair of classes $c_i, c_j$ such that either is not shared. \Cref{fig:exp1-separability} shows that our approach is consistently higher than the baseline, suggesting that the former can better avoid collisions from the merge. Surprisingly, the separability of the aggregated space is also higher than that of the end-to-end one, and it is higher for smaller numbers of shared classes. This again suggests that the task-specific models can better separate the samples and that the aggregated space can preserve this information. 

\paragraph{Summary}
Aggregating spaces of models trained on tasks sharing a portion of samples results in a space that is similar to the one of the end-to-end model but that allows significantly better classification performance and improved class separability, especially when task-specific models have fewer shared classes.

\subsection{Aggregating tasks sharing classes}\label{sec:exp-T2}
\paragraph{Setup}

\begin{wrapfigure}{r}{0.45\textwidth}
  \centering
  \includegraphics[width=0.44\textwidth]{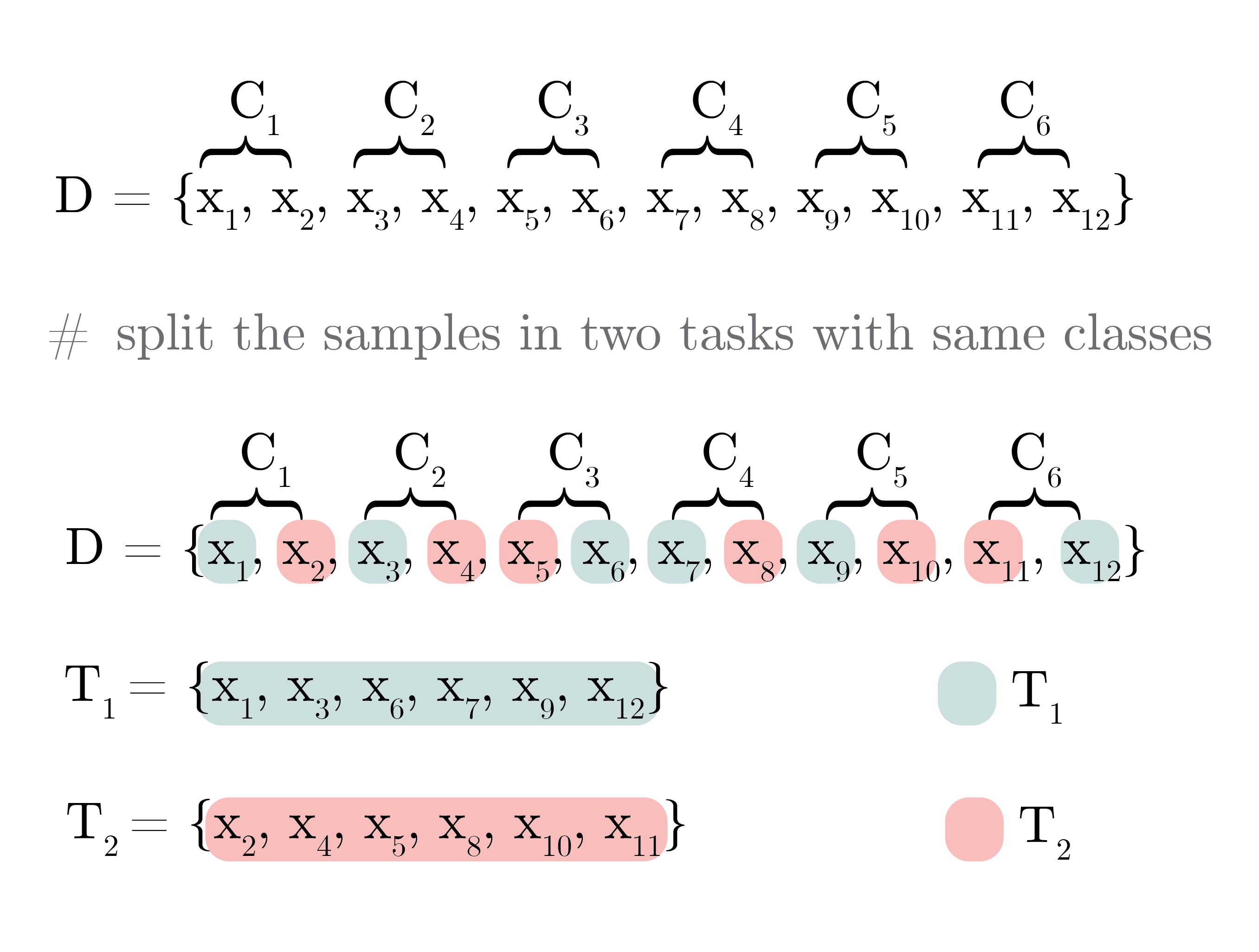} 
  \vspace{-0.5cm}
  \caption{Experiment outline.}\label{fig:exp2}
\end{wrapfigure}

In this experiment, we divide the dataset into two disjoint subsets, $A$ and $B$. Subset $A$ comprises 20\% samples labeled from the first half of the classes and 80\% instances labeled from the second one, while subset $B$ contains 80\% samples labeled from the first and 20\% samples labeled from the second. This arrangement ensures a distributional imbalance in terms of class labels between the two subsets.
Subsequently, we train two models on $A$ and $B$; each model learns to classify the distinct distribution of labels within its assigned subset. Post-training, we introduce an unseen set of anchor points to project the latent spaces of the two models to relative spaces. The final step involves unifying these relative spaces into a single latent space: similarly to the samples belonging to novel classes in \Cref{sec:exp-T1}, we only have one representation for each sample already in the unified latent space.

\begin{figure}
  \centering
  \includegraphics[width=\textwidth]{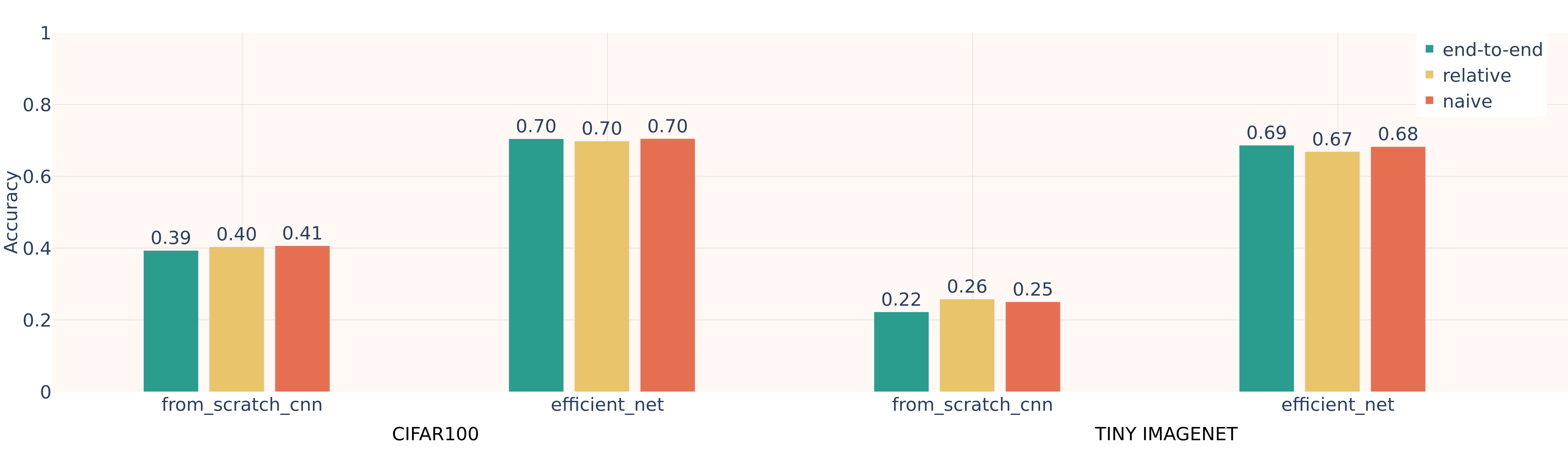}
  \vspace{-1cm}
  \caption{
    Classification analysis. Each barplot represents a different model: for each one; we compare its \emph{end-to-end} accuracy versus that of a classifier trained over the \emph{relative} aggregated space and that of a classifier trained over the \emph{naive} aggregated space.
  }
  \label{fig:same-classes-disj-samples}
\end{figure}

\paragraph{Results}
\Cref{fig:same-classes-disj-samples} shows that, in this case, relative aggregation does not improve over naive merging when considering classification accuracy. This suggests that a shared region of the latent space is required for the relative representations to be most effective. However, \cref{fig:exp2-separability} shows that the relative aggregated space still vastly outperforms the naive one regarding class separability. This last result hints at the relative aggregated space being better suited for classification, attributing the equality in results with the naive baseline to the expressiveness of the downstream classifier that can make up for the inferior separability of the naive space.

It is noteworthy to observe that the aggregated spaces, both the naive and the relative ones, can reach a comparable accuracy to the end-to-end model. This is a remarkable result, as it shows that the unified space can capture the information from both models, even though the two models were trained on disjoint tasks.
This is further confirmed by the similarity between original and relative aggregated being as high as $0.87, 0.83$ on \texttt{CIFAR100} for \texttt{EfficientNet} and \texttt{VanillaCNN} respectively, and $0.79$ and $0.84$ on \texttt{TinyImagenet}.

\paragraph{Summary} 
For sample-disjoint tasks sharing common classes but distinct distributions, the relative aggregated space effectively preserves information from both spaces, improving classification accuracy and class separability in downstream tasks.

\subsection{Aggregating totally disjoint tasks}\label{sec:exp-T3}
\begin{minipage}{0.6\textwidth}
  \paragraph{Setup} 
  Similarly to \cref{sec:exp-T2}, we split a dataset into two disjoint sets of samples, with the difference that these now also belong to disjoint class sets. 
    The first set comprises samples exclusively from the first half of the classes, while the second consists of samples from the remaining half. This ensures that the two subsets do not overlap in classes or instances, presenting an independent learning scenario for two models to be trained on. 
    Post-training, we introduce an unseen set of anchor points spanning all the classes, and use them to project the latent spaces into relative spaces.
\end{minipage}\hfill
\begin{minipage}{0.45\textwidth}
    \centering
    \includegraphics[width=\textwidth]{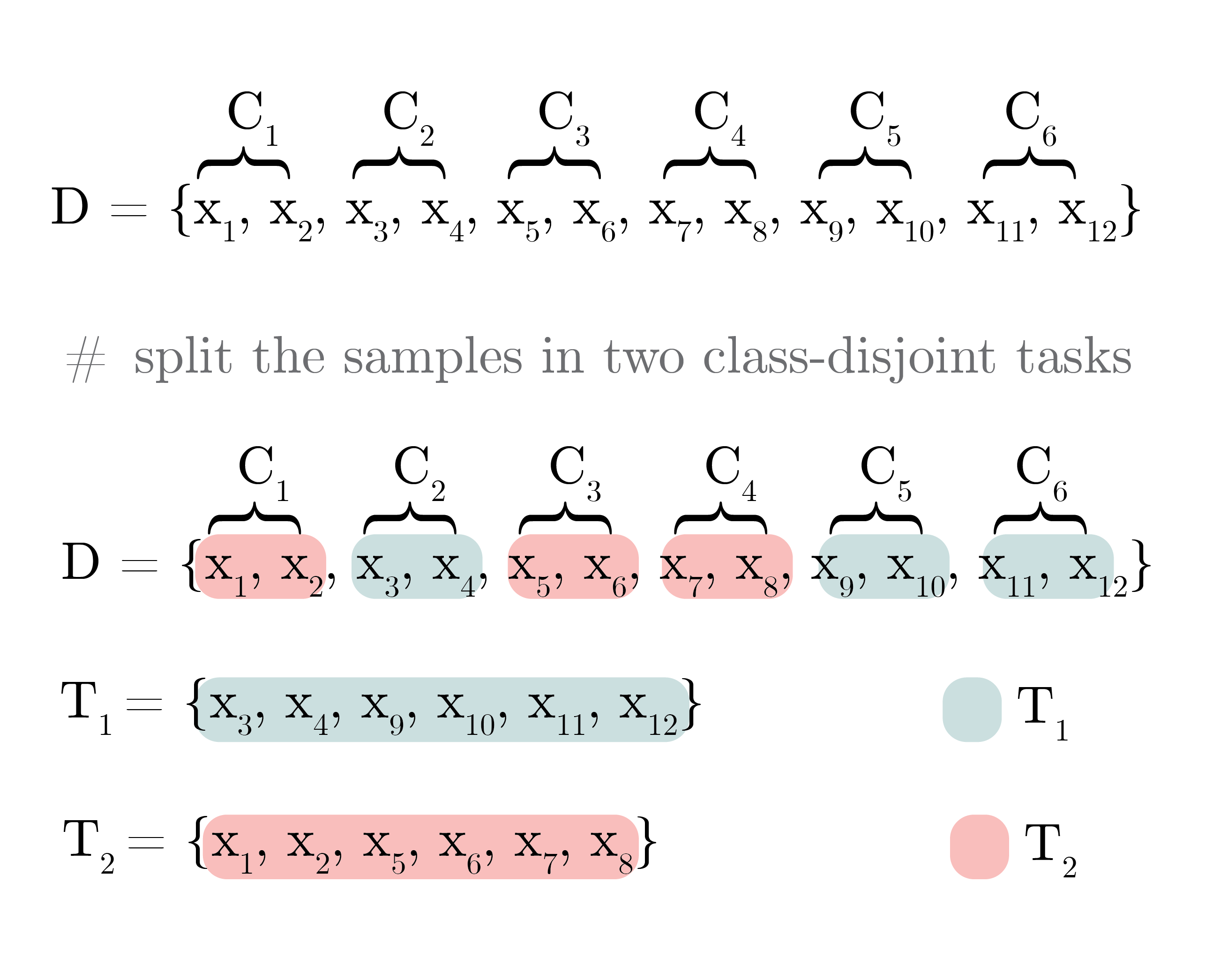} 
    \vspace{-1cm}
    \captionof{figure}{Experiment outline.}\label{fig:exp3}
\end{minipage}

\begin{figure}
        \centering
        \includegraphics[width=\textwidth]{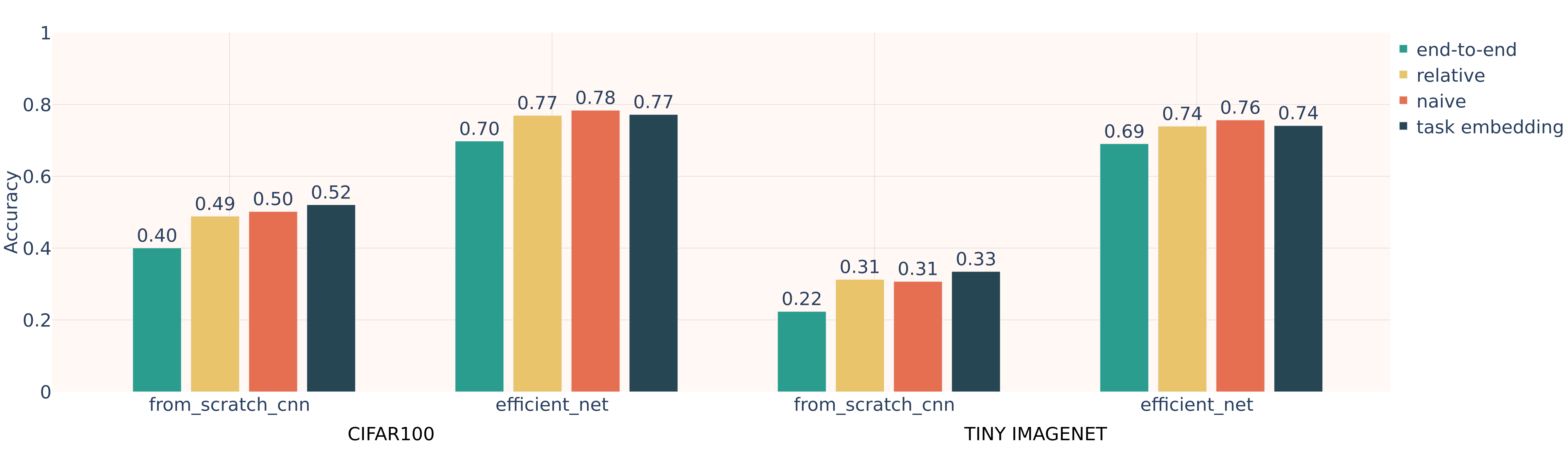}
        \vspace{-0.8cm}
        \caption{
        Classification analysis. Each barplot represents a different model: for each one; we compare its \emph{end-to-end} accuracy versus that of a classifier trained over the \emph{relative} aggregated space and that of a classifier trained over the \emph{naive} aggregated space.
        }
        \vspace{-0.5cm}
        \label{fig:totally-disjoint}
\end{figure}

\begin{figure}
        \centering
        \includegraphics[width=\textwidth]{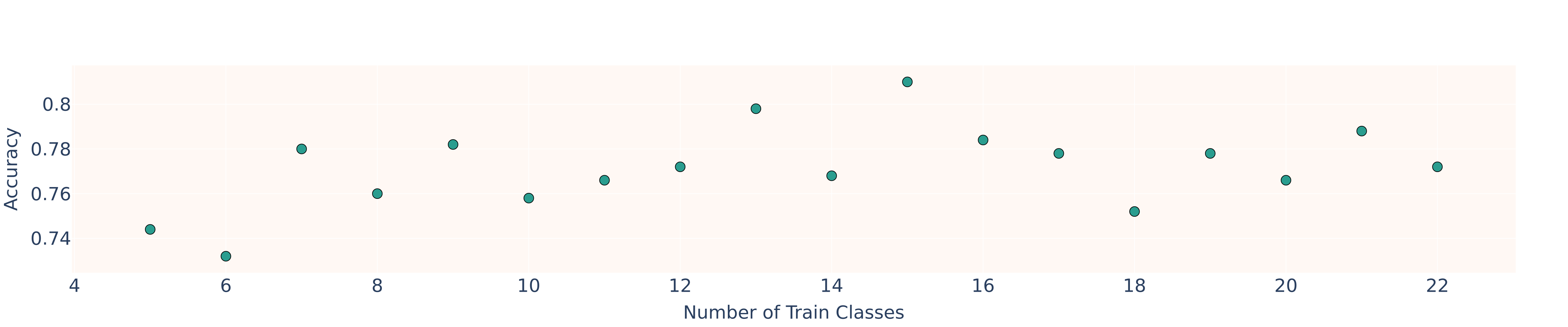}
        \vspace{-0.7cm}
        \caption{
        Classifier accuracy as the embedding space becomes less specific, where point $i$ represents the accuracy of a downstream classifier over classes $c_1, c_2, c_3$ embedded within a space containing classes $c_1, c_2, c_3, \dots, c_{i+3}$.
        }
        \label{fig:matrioska}
\end{figure}

In this way, each task-specific space will be projected with respect to anchors that are roughly half from the training distribution and half from out-of-distribution. Coming from disjoint tasks, each sample will then just be represented with its relative representation.
\paragraph{Results}
As the spaces to aggregate are class- and sample-disjoint, the similarity to the end-to-end space drops to $0.81, 0.83$ on \texttt{CIFAR100} and to \texttt{$0.72, 0.81$} on \texttt{TinyImageNet} for \texttt{EfficientNet} and \texttt{VanillaCNN} respectively. Interestingly, \cref{fig:totally-disjoint} shows that the space that allows us to obtain the best results is consistently the aggregated one, significantly outperforming the original space obtained by training a model end-to-end over the whole dataset. While surprising, this finding aligns with the increased performance over the tasks with fewer shared classes seen in \Cref{sec:exp-T1}. We envision two possibilities to explain this phenomenon.
%
%
%
\textbf{Hypothesis A}: 
The task-specific models can extract more specialized features for their classes, presenting the downstream classifier with better features.
To investigate this hypothesis, we start from the first three classes $C_1 = c_1, c_2, c_3$ of \texttt{CIFAR100} and gradually expand the set with a new class $c_i$ for $i=1, \dots, M$. We train a model on each class set $\{c_1, c_2, c_3\}, \dots, \{c_1, \dots, c_M\}$, use it to embed the samples with class in $C_1$ and then train a downstream classifier on this embedding space. Intuitively, the embeddings for three original classes will come from increasingly more crowded spaces with the addition of new classes. \Cref{fig:matrioska} shows that the accuracy has an increasing trend overall when adding new classes, therefore not supporting the hypothesis. Similarly, the second experiment in \cref{appendix:subclass-exp} compares the classification suitability of the embeddings of $M$ classes in a task-specific space versus the embeddings of the same classes in the space of all the $C$ classes. As before, the restriction over the  $M$ classes in the space of a model trained on all $C$ classes proves a better feature space than the space of a model that is only trained over the $M$ classes, again disconfirming the hypothesis.  \textbf{Hypothesis B}: the task-specific models trained on $C' \leq C$ classes leave a footprint in the embeddings, allowing the downstream classifier to discriminate among a smaller set of classes.
To verify whether this holds, we add a task-embedding layer to the downstream classifier and train it on the embedding space of the end-to-end model. The results, shown in \Cref{fig:totally-disjoint}, show that the accuracy rises to that of the aggregation techniques, confirming the hypothesis.
Intuitively, being trained on a smaller subset of classes, the task-specific embedders are better able to discriminate their class set. If the embeddings carry a footprint of the model, they also covertly reveal the task. The downstream classifier can exploit this information and discriminate among $C'$ classes instead of $C$, significantly lowering the complexity of the task and therefore increasing its accuracy.

\paragraph{Summary}
Relative aggregation allows merging sample- and class-disjoint spaces, granting better accuracy than an end-to-end model trained over the whole dataset due to task-specific footprints being embedded within the representations. 


\section{Conclusions}
In this work, we have studied the problem of aggregating spaces originating from different learning models, possibly more than one. We have seen how relative representations provide the means to perform this operation and analyzed various experimental settings spanning a set of varying assumptions. In particular, we have seen that the approach works best when a shared region of the space is available, and that it is still possible to obtain good results when this is not the case. We have studied the characteristics of the aggregated space in terms of representational power, separability, and similarity to an end-to-end space trained over the whole dataset, and have shown the benefits of the approach when compared to a naive merging baseline. In future work, we envision applying the approach to other settings, such as federated learning, where aggregating latent spaces involving anonymized anchors could provide an architecture-agnostic alternative to weight-space model merging.

\bibliography{references.bib}

\appendix

\section{Details}
We report here the technical details that are not present in the main manuscript.

\subsection{Dataset details}\label{appendix:dataset-details}
We employ two well-established benchmark datasets that offer a good trade-off between scale and tractability. In fact, the first experiment alone requires training and aggregating $70$ task-specific models.

\paragraph{CIFAR100}
\texttt{CIFAR-100} \citep{cifar100} is a popular dataset in the field of computer vision. It consists of 100 different classes, each containing 600 images, making a total of 60,000 labeled images for training and testing. \texttt{CIFAR-100} is widely used for benchmarking and evaluating the performance of machine learning and deep learning models in tasks such as image classification and object recognition. It offers a diverse range of objects and scenes, making it a challenging dataset for developing and testing algorithms in the field of image analysis and pattern recognition.

\paragraph{TinyImageNet}
\texttt{TinyImageNet} is a subset of the larger \texttt{ImageNet} dataset, focusing on a more manageable scale while maintaining its diversity. It consists of $100000$ images of 200 classes (500 for each class) downsized to $64\times 64$ colored images. Each class has $500$ training images, $50$ validation images and $50$ test images.

\subsection{Model details}
All our models are trained using the Adam optimizer from the \emph{torch.optim} suite.

\paragraph{VanillaCNN}
We utilized a vanilla \gls{cnn}. The model is defined by intermediate convolutional stages with $16$ and $32$ channels respectively, culminating in an embedding layer of 128 dimensions. The preprocessing pipeline only consists of normalization.

\paragraph{EfficientNet}
We employ \texttt{EfficientNet} as a pre-trained model for our experiments, specifically the \emph{efficientnet\_b0} variant provided by the \emph{TIMM} library.
Post the embedder, we employ a \gls{mlp} projector, which transitions from an input feature dimension of $1280$ to a hidden dimension of $256$, and ultimately to a projection dimension of $128$.
As for preprocessing, images are first converted to the PIL format, resized to $256$ pixels on the longer side, and then center-cropped to a consistent size of $224 \times 224$ pixels. Standard normalization ensures the input images are aptly conditioned for the model's requirements.

\subsection{Tools \& Technologies}
We use the following tools in all the experiments presented in this work:
\begin{itemize}
    \item \textit{PyTorch Lightning}, to ensure reproducible results while also getting a clean and modular codebase;
    \item \textit{NN-Template GrokAI (2021)}, to easily bootstrap the project and enforce best practices;
    \item \textit{Datasets by HuggingFace}, to access the datasets.
\end{itemize}

\section{Additional experiments}
We report here the results for the experiment configurations that were omitted in the manuscript, as well as experiments that were only mentioned.

\subsection{Qualitative visualization}

\begin{figure}[ht]
  \vspace{0.5cm}
  \centering
  \centering
  \begin{overpic}[width=.32\textwidth]{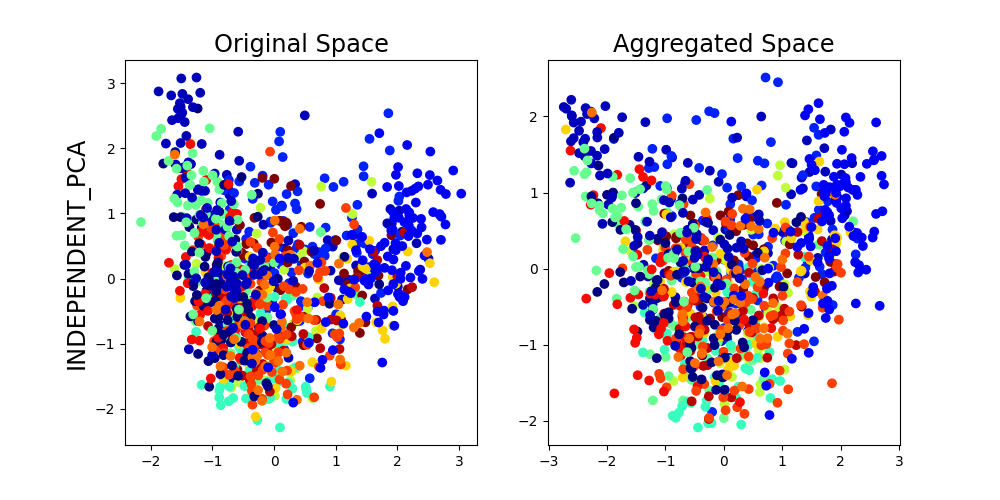}
    \put(40,50) {$S=80$}
    \put(-5, 10) {\rotatebox{90}{$N=5$}}
  \end{overpic}
  \hfill
  \centering
  \begin{overpic}[width=.32\textwidth]{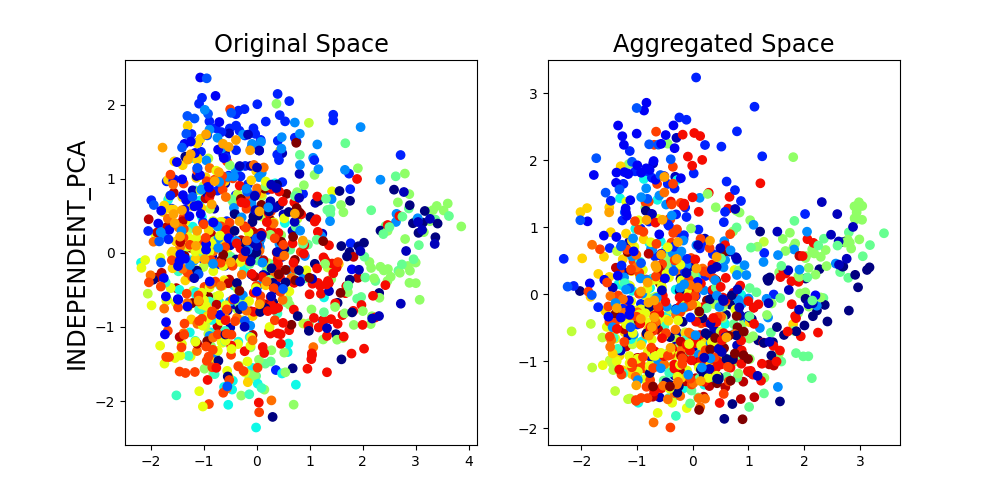}
    \put(40, 50) {$S=50$}
  \end{overpic}
  \hfill
  \centering
  \begin{overpic}[width=.32\textwidth]{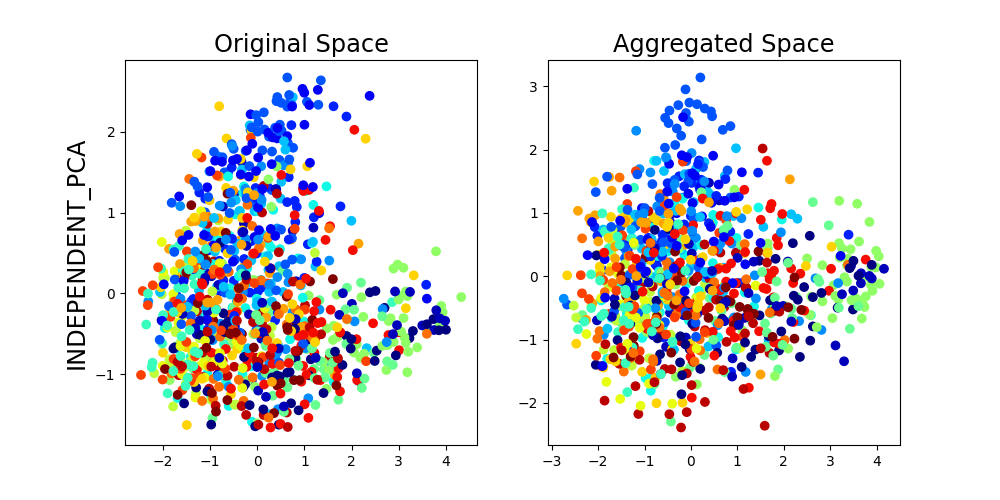}
    \put(40, 50) {$S=20$}
  \end{overpic}

  \vspace{1em}

  \centering
  \begin{overpic}[width=.32\textwidth]{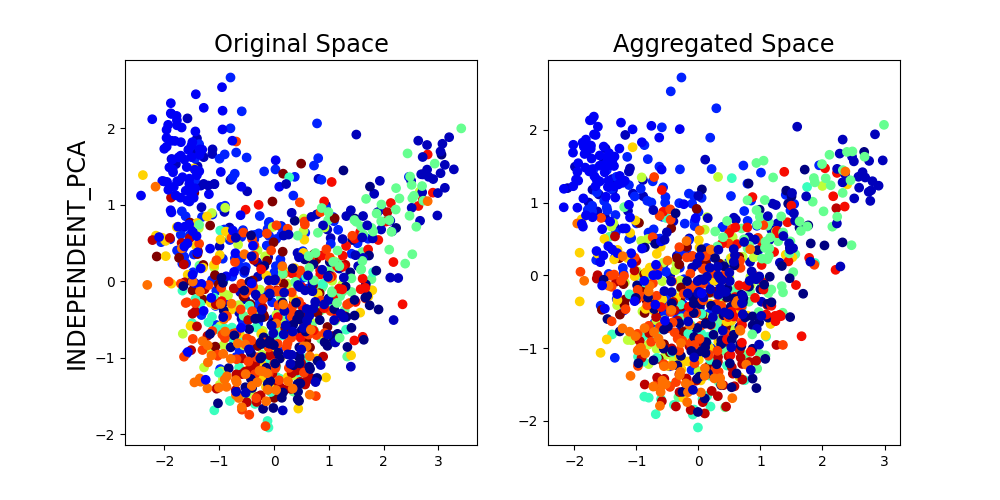}
    \put(-5, 10) {\rotatebox{90}{$N=10$}}
  \end{overpic}
  \hfill
  \centering
  \begin{overpic}[width=.32\textwidth]{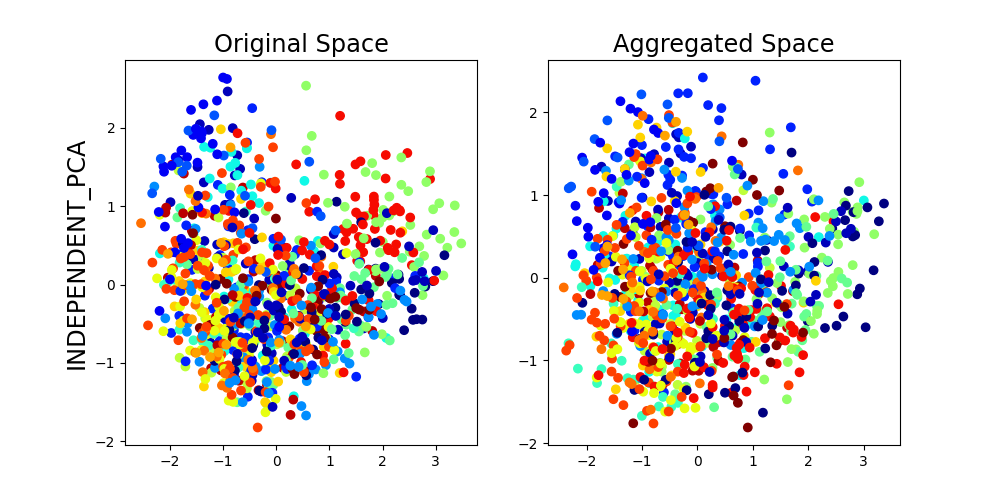}
  \end{overpic}
  \hfill
  \centering
  \begin{overpic}[width=.32\textwidth]{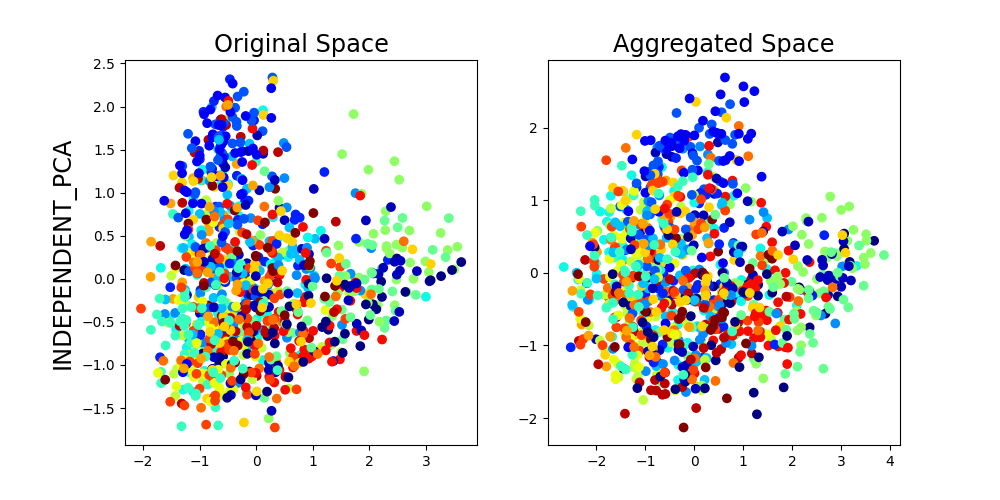}
  \end{overpic}

  \caption{
  \texttt{Experiment 1}. \gls{pca} visualizations of the latent space obtained training on the whole dataset versus aggregated latent spaces from models trained on partially disjoint tasks. Notably, the spaces remain similar despite reducing the shared core of classes.
  }\label{fig:part-shared-part-novel-pca}
\end{figure}

\Cref{fig:part-shared-part-novel-pca} compares the space of an end-to-end model trained on the whole dataset and the aggregated space from the task-specific models. Being the hardest to position, only the non-shared classes are visualized. 
Remarkably, even in this case, the aggregated spaces look similar to the one of the end-to-end model.

\subsection{Separability analysis}
\begin{figure}
        \centering
        \includegraphics[width=\textwidth]{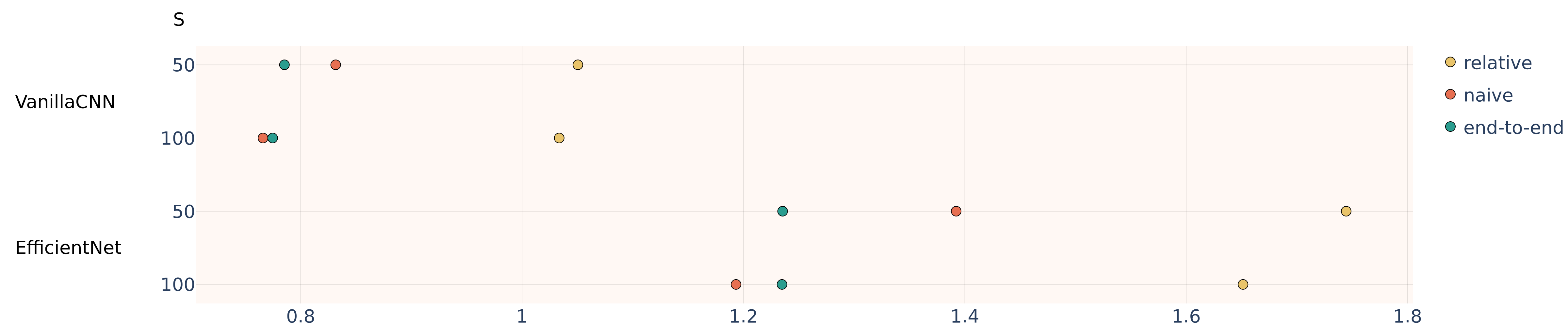}
        \caption{%
          Separability analysis over \texttt{TinyImageNet} for tasks sharing a sample portion: we compare the mean separability of the resulting space when using relative and naive aggregation for different values of $S$ and $N=25$, as well as the separability over the space of the end-to-end model trained over the whole dataset.
          The relatively aggregated space consistently results in the highest separability.
        }\label{fig:exp1-separability-tiny-N25}
\end{figure}

\begin{figure}
        \centering
        \includegraphics[width=\textwidth]{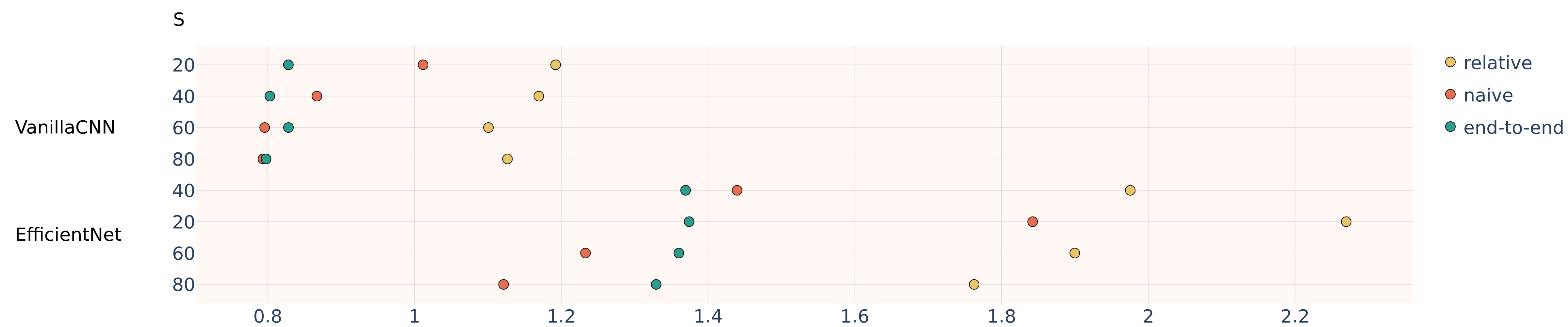}
        \caption{%
          Separability analysis over \texttt{CIFAR100} for tasks sharing a sample portion: we compare the mean separability of the resulting space when using relative and naive aggregation for different values of $S$ and $N=5$, as well as the separability over the space of the end-to-end model trained over the whole dataset.
          The relatively aggregated space consistently results in the highest separability.
        }\label{fig:exp1-separability-N5}
\end{figure}

\begin{figure}
        \centering
        \includegraphics[width=\textwidth]{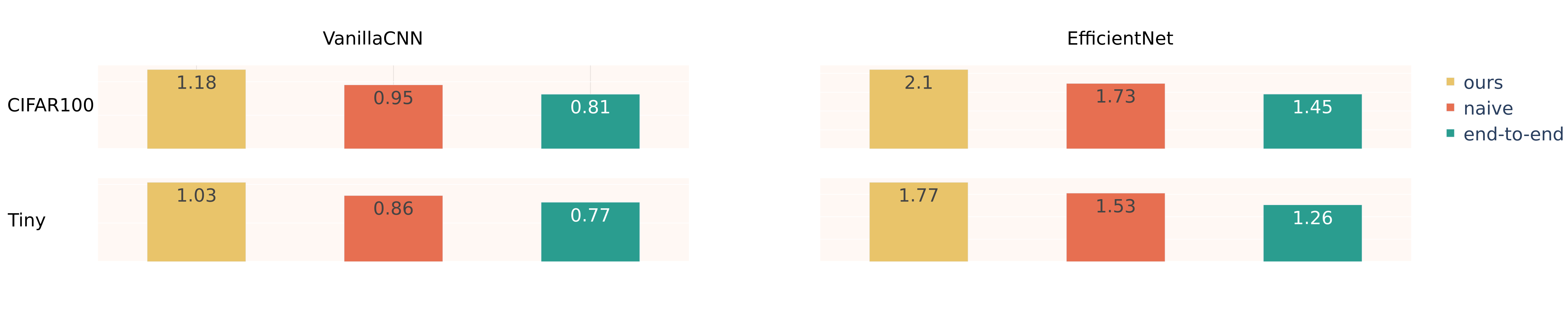}
        \caption{%
        Separability analysis over \texttt{CIFAR100} and \texttt{TinyImageNet} for sample-level disjoint tasks sharing the same class set: we plot the mean separability over the relatively aggregated space, the naively aggregated space, and the one of the end-to-end model. The relatively aggregated space consistently results in the highest separability.
        }\label{fig:exp2-separability}
\end{figure}

\begin{figure}
        \centering
        \includegraphics[width=\textwidth]{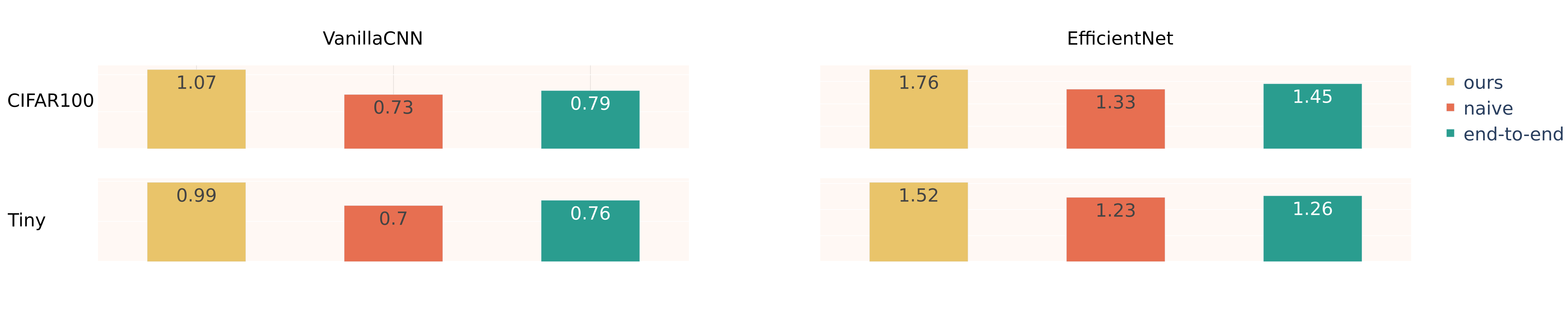}
        \caption{%
        Separability analysis over \texttt{CIFAR100} and \texttt{TinyImageNet} for tasks disjoint both at the sample and at the class level: we plot the mean separability over the relatively aggregated space, the naively aggregated space, and the one of the end-to-end model. The relatively aggregated space consistently results in the highest separability.%
        }\label{fig:exp3-separability}
\end{figure}

\Cref{fig:exp1-separability-N5} shows the separability for the spaces obtained in \cref{sec:exp-T1} for $N=5$. The same considerations given for $N=10$ apply. \Cref{fig:exp1-separability-tiny-N25} instead shows the same analysis for \texttt{TinyImageNet}. In \Cref{fig:exp2-separability}, we instead show the separability values for the aggregation of sample-level disjoint tasks, as described in \cref{sec:exp-T2}.
In \Cref{fig:exp3-separability} the same analysis is carried for spaces that are disjoint at the class and sample level, as described in \cref{sec:exp-T3}.

\subsection{Subclass experiment}\label{appendix:subclass-exp}
\begin{figure}
        \centering
        \includegraphics[width=\textwidth]{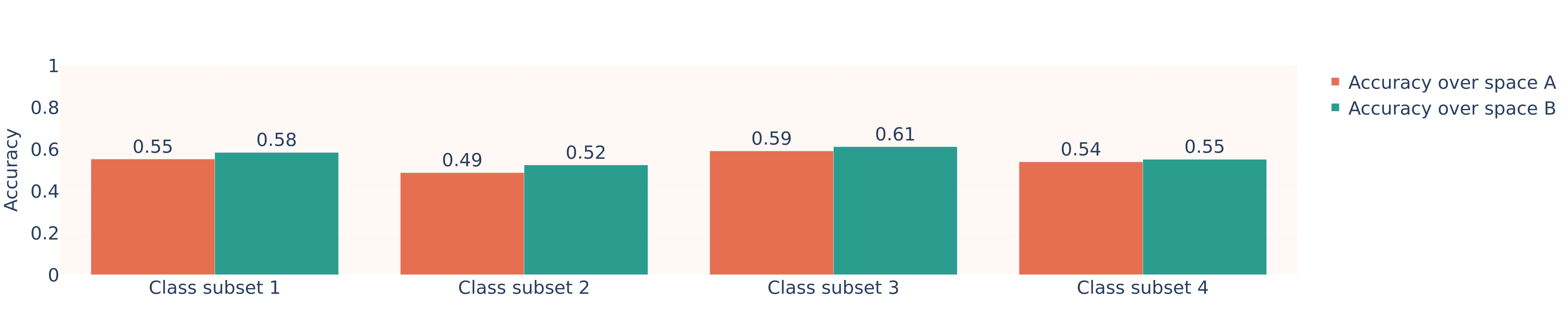}
        \caption{%
        Accuracy results for the experiment in \cref{appendix:subclass-exp}. Each barplot is a different task, space A is the feature space learned by the model trained on $M=30$ classes, and space B is that of the model trained on all the $C=100$ classes. A downstream classifier always obtain the best performance over the latter. 
        }\label{fig:class_subsets}
\end{figure}

In this experiment,  we compare the classification suitability of the embeddings of $M$ classes in a task-specific space versus the embeddings of the same classes in the space of all the $C$ classes. In practice, we randomly sample $M$ classes for each task, train a model on the corresponding sample subset, and then use it to embed an unseen test dataset for the same classes. We then also train a model from scratch over the whole dataset, and use it to embed the same sample subset (again, the samples for the selected $M$ classes). We then train two downstream classifiers to classify the sample label from the embeddings: one over the embeddings obtained by the task-specific model and one over the embeddings produced by the model trained on the whole dataset.  
We pick $M=30$ and $C=100$ for \texttt{CIFAR100}.
Intuitively, if the features learned by the task-specific model were better specialized, the classification accuracy of the downstream classifier would be higher than when trained over the features learned by the global model. We can see in \cref{fig:class_subsets} that this is not the case, as in any of the four attempts the features of the global model (space B) result in a greater accuracy.  
As before, the restriction over the  $M$ classes in the space of a model trained on all $C$ classes proves a better feature space than the space of a model that is only trained over the $M$ classes, again disconfirming hypothesis A in \cref{sec:exp-T3}.  

 \label{appendix:additional-exps}

\section{Background}
\subsection{Relative representations}
\begin{figure}
    \centering

    \begin{minipage}{0.51\linewidth}
        \centering
        \begin{tikzpicture}
            \begin{axis}[
                    axis lines=center,
                    view={60}{40}, 
                    domain=0:4,
                    y domain=0:4,
                    zmin=0, zmax=4,
                    enlargelimits=upper,
                    colormap/viridis,
                    samples=50
                ]

                \addplot3 [surf, opacity=0.5]
                {sin(deg(sqrt(x^2+y^2))) + 2};

                \addplot3 [only marks, mark=*, red] coordinates {
                        (1,1,1.2) 
                        (2,2,0.5) 
                        (3,3,2.5) 
                    };

                \addplot3 [only marks, mark=*, blue] coordinates {
                        (1.5,1.5,1.7) 
                    };

                \node[coordinate,pin=left:{\(a_1\)}] at (axis cs:1,1,1.2) {};
                \node[coordinate,pin=right:{\(a_2\)}] at (axis cs:2,2,0.5) {};
                \node[coordinate,pin=above:{\(a_3\)}] at (axis cs:3,3,2.5) {};
                \node[coordinate,pin=below:{\(x\)}] at (axis cs:1.5,1.5,1.7) {};

            \end{axis}
        \end{tikzpicture}
    \end{minipage}
    \hfill
    \begin{minipage}{0.48\linewidth}
        \centering
        \tdplotsetmaincoords{60}{110} 
        \begin{tikzpicture}[tdplot_main_coords]
            \begin{axis}[
                    axis lines=center,
                    xlabel={$d(x, a_1)$},
                    ylabel={$d(x, a_2)$},
                    zlabel={$d(x, a_3)$},
                    every axis x label/.style={
                            at={(ticklabel* cs:1.05)},
                            anchor=west,
                        },
                    every axis y label/.style={
                            at={(ticklabel* cs:1.05)},
                            anchor=west,
                        },
                    every axis z label/.style={
                            at={(ticklabel* cs:1.15)},
                            anchor=west,
                        },
                    xtick=\empty,
                    ytick=\empty,
                    ztick=\empty,
                    xmin=0, xmax=4,
                    ymin=0, ymax=4,
                    zmin=0, zmax=4
                ]

                \addplot3 [only marks, mark=*, mark options={color=blue}] coordinates {(1,2,3)};

                \addplot3 [dotted, thick] coordinates {(0,2,3) (1,2,3)};
                \addplot3 [dotted, thick] coordinates {(1,0,3) (1,2,3)};
                \addplot3 [dotted, thick] coordinates {(1,2,0) (1,2,3)};
                \addplot3 [dotted, thick] coordinates {(0,0,3) (0,2,3)};
                \addplot3 [dotted, thick] coordinates {(0,2,0) (0,2,3)};
                \addplot3 [dotted, thick] coordinates {(0,0,3) (1,0,3)};
                \addplot3 [dotted, thick] coordinates {(1,0,0) (1,0,3)};
                \addplot3 [dotted, thick] coordinates {(1,2,0) (1,0,0)};
                \addplot3 [dotted, thick] coordinates {(0,2,0) (1,2,0)};

                \addplot3 [only marks, mark=*, mark options={color=red}] coordinates {
                        (0,2,3) 
                        (1,0,3) 
                        (1,2,0) 
                    };

            \end{axis}

        \end{tikzpicture}
    \end{minipage}
    \caption{Relative representations: (left) a sample $x$ and three anchor samples $a_1, a_2, a_3$ are embedded in a latent space. (right) the new representation of $x$ is given by its distance with respect to the anchors.}\label{fig:rel-reps}
\end{figure}
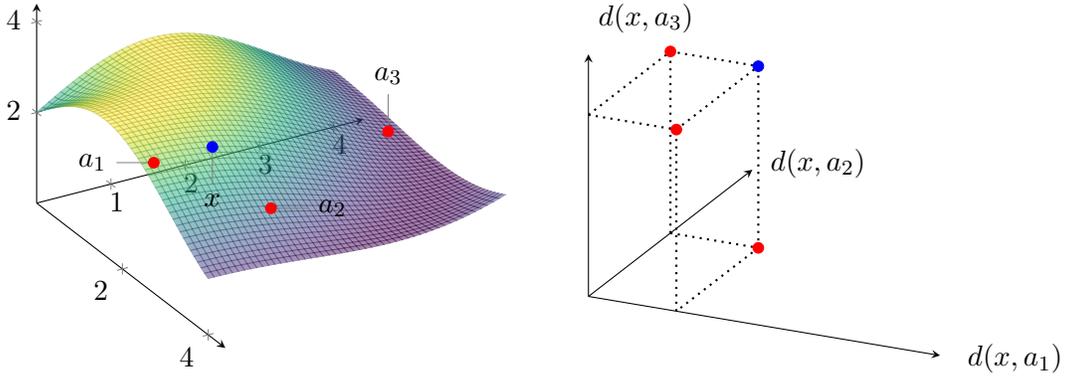

Relative representations encode data points as distances with respect to a set of training points termed anchors. \cite{moschella2023relative} empirically show that in many scenarios two spaces become aligned when passing to this distance-based representation, claiming that stochasticities in the training process often result in an angle-preserving transformation between the two latent spaces. 
\Cref{fig:rel-reps} shows a visual representation of the relative representation framework.

\subsection{Similarity measures}
The most common representational similarity measure is \gls{cka} \citep{Kornblith2019-vr}, that is defined as:
\begin{equation}
    \label{eq:cka}
    m_{\operatorname{CKA}}(\R,\Rprime) = \frac{\operatorname{HSIC}(\Sm, \Smp)}{\sqrt{\operatorname{HSIC}(\Sm, \Sm) \mathrm{HSIC}(\Smp, \Smp)}},
\end{equation}
where $\operatorname{HSIC}(\Sm, \Smp) = \frac{1}{(N-1)^2} \operatorname{tr}(\bm{SHS'H})$, $\bm{H} = \bm{I} - \frac{1}{N} \bm{1}\bm{1}^\transp$ is a centering matrix, and $\bm{1}$ is a vector of $N$ ones.
The denominator is introduced to scale \gls{cka} between 0 and 1, with a value of $1$ indicating equivalent representations. We employ the linear kernel version, as the difference with more complex kernels is usually negligible \citep{Kornblith2019-vr}.

Most notably, \gls{cka} is invariant to orthogonal transformations and isotropic scaling, assuming invariant similarity measures for RSM computation. It is worth remarking that relative representations are invariant to angle-preserving transformations, a superclass of the orthogonal transformations.

\subsection{Separability measures}\label{appendix:separability}
Let's denote the aggregated space as \( \mathbf{X}_{\text{aggregated}} \). For simplicity, consider a scenario where the aggregated space consists of samples from two different classes \( C_1 \) and \( C_2 \). The goal is to assess the separability of these two classes in the aggregated space. To do so, we first calculate the centroid (mean vector) for each class in the aggregated space:
\begin{align*}
    \mathbf{c}_{C_1} & = \frac{1}{|C_1|} \sum_{x \in C_1} \mathbf{x} \\
    \mathbf{c}_{C_2} & = \frac{1}{|C_2|} \sum_{x \in C_2} \mathbf{x}
\end{align*}
Then, we calculate the Euclidean distance between the centroids of the two classes:
\[
    d_{\text{inter}} = \|\mathbf{c}_{C_1} - \mathbf{c}_{C_2}\|
\]
as well as the average distance of samples in each class to their respective centroid:
\begin{align*}
    d_{\text{intra}_{C_1}} & = \frac{1}{|C_1|} \sum_{x \in C_1} \|\mathbf{x} - \mathbf{c}_{C_1}\| \\
    d_{\text{intra}_{C_2}} & = \frac{1}{|C_2|} \sum_{x \in C_2} \|\mathbf{x} - \mathbf{c}_{C_2}\|
\end{align*}
A separability score can then be defined as the ratio of inter-class distance to the average intra-class distance:
\[
    S = \frac{d_{\text{inter}}}{\frac{1}{2} (d_{\text{intra}_{C_1}} + d_{\text{intra}_{C_2}})}
\]
Intuitively, an high value of $S$ indicates that the classes are well-separated in the aggregated space, meaning there is little to no collision, while a value that is close to or less than $1$  indicates that there is a significant overlap or collision of classes in the aggregated space.
For a set of classes, we generalize the above approach by calculating pairwise separability scores for all combinations of class pairs and then taking the average or the minimum of these scores as an overall indicator.

\end{document}